%% file: emnlp2020.tex
\documentclass[11pt,a4paper]{article}
\usepackage[hyperref]{emnlp2020}
\usepackage{times}
\usepackage{latexsym}

\usepackage{subcaption}
\usepackage{microtype}

\aclfinalcopy 

\usepackage{booktabs}
\input{math_commands.tex}

\usepackage{comment}

\usepackage{algorithm}
\usepackage{hyperref}
\usepackage{url}
\usepackage{todonotes}
\usepackage{color,soul}
\usepackage{graphicx}
\usepackage[normalem]{ulem}
\usepackage{mathtools}
\usepackage{algorithmicx}
\usepackage{algpseudocode}
\usepackage{amsmath,amssymb}

\newcommand{\reals}{\mathbb{R}}

\newcommand{\zz}{\mathbf{z}}

\newcommand{\modelM}{\mathbb{M}}

\usepackage{url}

\title{Analyzing Individual Neurons in Pre-trained Language Models}

\author{
Nadir Durrani ~~~ Hassan Sajjad ~~~ Fahim Dalvi ~~~ Yonatan Belinkov\textsuperscript{*} \\ 
{\tt \{ndurrani,hsajjad,faimaduddin\}@hbku.edu.qa} \\ 
Qatar Computing Research Institute, HBKU Research Complex, Doha 5825, Qatar \\\\ 
\textsuperscript{*}MIT Computer Science and Artificial Intelligence Laboratory and Harvard \\  
 John A.\ Paulson School of Engineering and Applied Sciences, Cambridge, MA, USA \\ 
{\tt belinkov@csail.mit.edu}
}

\begin{document}
\maketitle

\begin{abstract}

While a lot of analysis has been carried to demonstrate linguistic knowledge captured by the representations learned within
deep NLP models, very little attention has been paid towards individual neurons. We carry out 
a neuron-level analysis 
using core linguistic tasks of predicting morphology, syntax and semantics, on pre-trained language models, 
with questions like: 
 i) do individual neurons in pretrained models capture linguistic information? ii) which parts of the network learn more about certain linguistic phenomena? iii) how distributed or focused is the information? and iv) how do various architectures differ in learning these properties?
We found 
small subsets of neurons to predict linguistic tasks, with lower level tasks (such as morphology) localized in fewer neurons, compared to higher level task of predicting syntax. Our study reveals interesting cross architectural comparisons. 
For example, we found neurons in XLNet to 
be more localized and disjoint when predicting properties compared to BERT and others, where they are more distributed and coupled. 

\end{abstract}

\section{Introduction}


Transformer-based neural language models have constantly pushed the state-of-the-art 
in downstream NLP tasks such as \emph{Question Answering}, \emph{Textual Entailment}, etc. \cite{rajpurkar-etal-2016-squad, wang-etal-2018-glue}. 
Central to 
this revolution 
is the contextualized embedding, where each word is assigned 
a vector based on the entire input sequence, 
allowing it to capture not only 
a static semantic meaning but also a contextualized meaning. 

Previous work on analyzing neural networks showed that while learning rich NLP tasks such as machine translation and language modeling, these deep models capture fundamental linguistic phenomena such as word morphology, syntax and various other relevant properties of interest \cite{shi-padhi-knight:2016:EMNLP2016,adi2016fine,belinkov:2017:acl, belinkov:2017:ijcnlp, dalvi:2017:ijcnlp, blevins-etal-2018-deep}.


\noindent More recently \newcite{liu-etal-2019-linguistic} and \newcite{tenney2018what} used probing classifiers to analyze 
pre-trained neural language models on a variety of 
sequence labeling tasks and demonstrated that contextualized representations encode useful, transferable features of language. 
While 
most of the previous studies emphasize and analyze representations as a whole, 
very little work has been carried to analyze individual neurons 
in deep NLP models.

Studying 
individual neurons can facilitate understanding of the inner workings of neural networks 
\cite{karpathy2015visualizing,dalvi:2019:AAAI, suau2020finding} and have other potential benefits such as controlling bias and manipulating system's behaviour \cite{bau2018identifying}, model distillation and compression \cite{rethmeier2019txray}, efficient feature selection \cite{dalvi-2020-CCFS}, and guiding architectural search.


In this work, we put the representations learned within pre-trained transformer models under the microscope and carry out a 
fine-grained neuron level analysis
with respect to various 
linguistic properties. 
We target questions such as: i) do individual neurons in pretrained models capture linguistic information? ii) which parts of the network learn more about certain linguistic phenomena? iii) how distributed or focused is the information? and iv) how do various architectures differ in learning these properties?  


A typical methodology 
in previous work on analyzing representations 
trains probing classifiers 
using the representations learned within a neural model, 
to predict 
the understudied task. We also use a probing classifier approach to analyze individual neurons. Since neurons are multivariate in nature and 
work in groups, we additionally use elastic-net regularization that encourages individual and group of neurons to play a role in the training of the classifier. Given a trained classifier, we consider the weights assigned to each neuron as a measure of their importance with respect to the understudied linguistic task. We use probes with high \emph{selectivity}  \cite{hewitt-liang-2019-designing} to ensure that our results reflect the property of representations and not the probe's capacity to learn.


We choose 4 pre-trained models: ELMo \cite{peters-etal-2018-deep}, its transformer variant T-ELMo \cite{peters-etal-2018-dissecting}, 
BERT \cite{devlin-etal-2019-bert} and XLNet \cite{yang2019xlnet} --
covering a varied set of modeling choices, including the building blocks (recurrent networks versus Transformers), optimization objective (auto-regressive versus non-autoregressive), and model depth and width. 
Our cross architectural analysis yields the following insights:
%
%
\begin{itemize}
    \item 
    Information across networks is distributed, but it is possible to extract a very small subset of neurons to predict a 
    linguistic task with the same accuracy as using the entire network.
    \item Low level tasks such as predicting morphology 
    require fewer neurons compared to high level tasks such as predicting syntax.
    \item Some phenomena (e.g. \emph{Verbs}) are distributed across 
    many neurons while others (e.g. \emph{Interjections}) are localized 
    in a 
    fewer neurons.
    \item  Lower layers contain more word-level specialized 
    neurons, 
    and higher layers contain neurons specialized in syntax-level information. 
    \item BERT is the most distributed model with respect to all properties while XLNet exhibits focus 
    with the most disjoint set of neurons and layers designated for different linguistic properties. 
\end{itemize}






\section{Methodology}
\label{sec:methodology}


A common approach for probing neural network components against linguistic properties is to train a linear 
classifier using the 
activations generated from the trained neural network as static features. 
The underlying 
assumption is that if a simple linear model can predict a linguistic property, then the representations implicitly encode this information.

\paragraph{Probe:} We go a level deeper 
and identify 
neurons within the learned representations to carry out a more fine-grained neuron\footnote{In our terminology, a neuron is one dimension in a high-dimensional representation, even when the representation is the output of a complex operation such as a transformer block. 
} level analysis.  
We use a logistic regression classifier with elastic-net regularization \cite{Zou05regularizationand}. 
The weights of the 
trained classifier 
serve as a proxy to select the most relevant features\footnote{We use features and neurons interchangeably in the paper.} within the learned representations, to predict a linguistic property. 
Formally, consider a pre-trained neural language model $\mathbf{M}$ with $L$ layers: $\{l_1, l_2, \ldots, l_L\}$. Given a dataset $\sD=\{w_1, w_2, ..., w_N\}$ 
with a corresponding set of linguistic annotations $\sT=\{t_{w_1}, t_{w_2}, ..., t_{w_N}\}$, we map each word $w_i$ in the data $\sD$ to a sequence of latent representations: $\sD\xmapsto{\modelM}\zz = \{\zz_1, \dots, \zz_n\}$. The representations can either be extracted from the entire model 
or just from an individual layer. The model is trained by minimizing the following loss function:
%
\begin{equation}
\mathcal{L}(\theta) = -\sum_i \log P_{\theta}(t_{w_i} | w_i) + \lambda_1 \|\theta\|_1 + \lambda_2 \|\theta\|^2_2 \nonumber
\end{equation}
%
where $P_{\theta}(t_{w_i} | w_i)$
 is the probability that word $i$ is assigned property $t_{w_i}$. The weights \mbox{$\theta \in \reals^{D \times T}$} are learned with gradient descent. Here $D$ is the dimensionality of the latent representations $\zz_i$ and $T$ is the number of tags (properties) in the linguistic tag set, which the classifier is predicting. The terms $\lambda_1 \|\theta\|_1$ and $\lambda_2 \|\theta\|^2_2$ correspond to $L1$ and $L2$ regularization. This combination, known as 
elastic-net, 
strikes a balance between identifying very focused localized features ($L1$) versus  distributed neurons ($L2$). We use a grid search algorithm described in \textbf{Search}, to find the most appropriate set of lambda values. But let us describe the neuron ranking algorithm first.

\paragraph{Neuron Ranking Algorithm:}


Once 
the classifier has been trained, 
our goal is to retrieve individual or a group of neurons (some subset of features of the latent representation) that are the most relevant for predicting a particular linguistic property $\sT$ of interest. We use the neuron ranking algorithm as described in \newcite{dalvi:2019:AAAI}. Given the trained classifier \mbox{$\theta \in \reals^{D \times T}$}, the algorithm extracts a ranking of the $D$ neurons in the model $\modelM$. For each label\footnote{We use label and sub-property interchangeably.} $t$ 
in task $\sT$, the weights are sorted by their absolute values in descending order. To select $N$ most salient neurons w.r.t. the task $\sT$, an iterative process is carried. The algorithm starts with a small percentage of the total weight mass and selects the most salient neurons for each 
sub-property (e.g. \emph{Nouns} in POS tagging) until the set reaches the specified size $N$. 

\paragraph{Search:} 

The search criteria is driven through ablation of weights in the trained classifier. Once the classifier is trained, we select $M$\footnote{$M$ is set to 20\% of the network in our experiments} top and bottom features according to our ranked list (obtained using neuron ranking algorithm described above) and zero-out the remaining features. We then compute score for each lambda set ($\lambda_1$, $\lambda_2$) as:
\begin{equation}
\mathcal{S}(\lambda_1, \lambda_2) =  \alpha ( A_t - A_b) - \beta (A_z - A_l) \nonumber
\end{equation}
where $A_t$ is the accuracy of the classifier retaining top neurons and masking the rest, $A_b$ is the accuracy retaining bottom neurons, $A_z$ is the accuracy of the classifier trained using all neurons but without regularization, and $A_l$ is the accuracy with the current lambda set. The first term ensures that we select a lambda set where accuracies of top and bottom neurons are further apart and the second term ensures that we prefer weights that incur a minimal loss in classifier accuracy due to regularization.\footnote{For some lambdas, for example  
with high value of L1, the classifier prefers sparsity, i.e. selects fewer very focused neurons but performs very badly on the task.} We set $\alpha$ and $\beta$ to be $0.5$ in our experiments. This formulation enables the search to be automated, compared to \newcite{dalvi:2019:AAAI} where the lambdas were selected manually, which we found to be cumbersome and error-prone.

\paragraph{Minimal Neuron Selection:}  Once we have obtained the best regularization lambdas, we follow a 3-step process to extract minimal neurons for any downstream task: i) train a classifier to predict the task using all the neurons (call it \emph{Oracle}), ii) obtain a neuron ranking based on the ranking algorithm described above,  iii) choose the top $N$ neurons from the ranked list and retrain a classifier using these, iv) repeat step 3 by increasing the size of $N$,\footnote{We increment by adding 1\% neuron at every step.} until the classifier obtains an accuracy close (not less than a specified threshold $\delta$) to the \emph{Oracle}. 


\paragraph{Control Tasks:} While there is a plethora of work demonstrating that contextualized representations encode a continuous analogue of discrete linguistic information, a question has also been raised recently if the representations actually encode linguistic structure or whether the probe memorizes the understudied task.  We use \emph{Selectivity} as a criterion to put a ``linguistic task's accuracy in context with the probe’s capacity to memorize from word types'' \cite{hewitt-liang-2019-designing}. It is defined as the difference between linguistic task accuracy and control task accuracy. An effective probe is recommended to achieve high linguistic task accuracy and low control task accuracy. 
The control tasks for our probing classifiers are defined by mapping each word type $x_i$ to a randomly sampled behavior $C(x_i)$, from a set of numbers \{$1 \dots T$\} where $T$ is the size of tag set to be predicted in the linguistic task. 
The sampling is done using the empirical token distribution of the linguistic task, so the marginal probability of each label is similar. We compute \emph{Selectivity} by training classifiers using all and the selected neurons.

\section{Experimental Setup}
\label{sec:expSetup}

\paragraph{Pre-trained Neural Language Models:} We present results with 4 pre-trained models: 
ELMo \cite{peters-etal-2018-deep}, 
and 3 transformer architectures: Transformer-ELMo \cite{peters-etal-2018-dissecting}, BERT \cite{devlin-etal-2019-bert} and XLNet \cite{yang2019xlnet}. The ELMo model is trained using a bidirectional recurrent neural network (RNN) with 3 layers each of size 1024 dimensions. Its transformer equivalent (T-ELMo) is trained with 7 layers but with the same hidden layer size. 
The BERT model is trained as an auto-encoder with a 
dual objective function of predicting 
masked words and next sentence in auto-encoding fashion. We use base version (13 layers and 768 dimensions). Lastly we included XLNet-base which is trained with the same parameter settings (number and size of hidden layers) as BERT, but with a permutation based auto-regressive objective function. 

\paragraph{Language Tasks:} We evaluated our method on 4 linguistic tasks: POS-tagging using 
the Penn TreeBank \cite{marcus-etal-1993-building}, syntax tagging (CCG supertagging)\footnote{CCG captures global syntactic information locally at the word level by assigning a label to each word annotating its syntactic role in the sentence. The annotations can be thought of as a function that takes and return syntactic categories (like an NP: Noun phase).} using CCGBank \cite{hockenmaier2006creating}, syntactic chunking using CoNLL 2000 shared task dataset \cite{tjong-kim-sang-buchholz-2000-introduction}, and semantic tagging using the Parallel Meaning Bank data \cite{abzianidze-EtAl:2017:EACLshort}. We used standard splits for training, development and test data (See Appendix \ref{subsec:data}) 

\paragraph{Classifier Settings:} 

We used linear probing classifier with elastic-net regularization, 
using a categorical cross-entropy loss, optimized by Adam \cite{kingma2014adam}. Training is run with shuffled mini-batches of size 512 and stopped after 10 epochs. The regularization weights are trained using grid-search algorithm.\footnote{See Appendix \ref{subsec:hyperparameters} for hyperparameters selected for each task.} For sub-word based models, we use the last activation value to be the representative of the word as prescribed for the embeddings extracted from Neural MT models \cite{durrani-etal-2019-one} and pre-trained Language Models \cite{liu-etal-2019-linguistic}. Linear classifiers are a popular choice in analyzing deep NLP models due to their better interpretability \cite{qian-qiu-huang:2016:EMNLP2016,belinkov-etal-2020-analysis}. \newcite{hewitt-liang-2019-designing} have also shown linear probes to have higher \emph{Selectivity}, a property deemed desirable for more interpretable probes.  Linear probes are particularly important for our method as we use the learned weights as a proxy to measure the importance of each neuron. 


\section{Evaluation}
\label{sec:eval}

\subsection{Ablation Study} 

First we evaluate our 
rankings as obtained by the 
neuron selection algorithm presented in Section \ref{sec:methodology}. We extract a ranked list of neurons with respect to each property set (linguistic task $T$) and ablate neurons in the classifier to verify the rankings. This is done by \emph{zeroing-out} all the activations in the test, except for the selected $M\%$ neurons. We select top, random and bottom 20\%\footnote{The choice of 20\% is arbitrary. We did not experiment much with it as this was merely to select best lambdas and to demonstrate the efficacy of rankings.} neurons to evaluate our rankings. Table \ref{tab:classifier_ablation_mask_out} shows the efficacy of our rankings, with low performance (prediction accuracy) 
using only the bottom or random neurons versus using only the top neurons. 
The accuracy of random neurons is 
high in some cases (for example CCG, a task related to predicting syntax) showing when the underlying task is complex, 
the information related to it is more distributed 
across the network causing redundancy. 

\begin{table}[t]									
\centering					
\footnotesize
    \begin{tabular}{l|cccc}									
    \toprule									
    & BERT & XLNet & T-ELMo & ELMo \\		
    \midrule
    \multicolumn{5}{c}{POS}  \\
    \toprule
    All & 96.04 & 96.13 & 96.39 & 96.48 \\
    Top & 90.16 & 92.28 & 91.96 & 83.01 \\
    Random & 28.45 & 58.17 & 48.40 & 30.80 \\
    Bottom & 16.86 & 44.64 & 21.11 & 15.56 \\
    \midrule 
    \multicolumn{5}{c}{SEM}  \\
    \midrule
    All & 92.09  & 92.64 & 91.94 & 93.29 \\
    Top & 84.32 & 90.70 & 84.16 &  81.23\\
    Random & 64.28 & 72.14 & 66.15 & 75. 82\\
    Bottom & 59.02 & 25.37 & 36.14 & 58.32 \\
    \midrule 
    \multicolumn{5}{c}{Chunking}  \\
    \midrule
    All & 95.01 & 94.15 & 93.43 & 93.14 \\
    Top & 89.01 & 89.16 & 87.63 & 82.51 \\
    Random & 75.83 & 75.26 & 79.40 & 70.23 \\
    Bottom & 66.82 & 46.66 & 48.11 & 64.39 \\
    \midrule 
    \multicolumn{5}{c}{CCG}  \\
    \toprule
    All & 92.16  & 92.55 & 91.70 & 91.19 \\
    Top & 75.13 & 76.48 & 71.31 & 68.19 \\
    Random & 71.11 & 63.71 & 68.23 & 41.17\\
    Bottom & 59.13 & 62.42 & 67.11 & 30.32  \\
    \bottomrule
    \end{tabular}
    \caption{Ablation Study: Selecting all, top, random and bottom 20\% neurons and zeroing-out remaining to evaluate classifier accuracy on blind test (averaged over 3 runs). See Appendix \ref{subsec:ablationStudy} for dev results.}

\label{tab:classifier_ablation_mask_out}						    
\end{table}

\begin{table}[t]									
\centering		
\resizebox{\columnwidth}{!}{
\begin{tabular}{l|cccc}									
\toprule									
 & BERT & XLNet & T-ELMo & ELMo \\		
 \midrule
 Neu$_a$ & 9984  & 9984 & 7168 & 3072 \\
\midrule
\multicolumn{5}{c}{POS}  \\
\midrule
Neu$_t$ & 400/4\%  & 400/4\% & 430/6\% & 368/12\%\\
Acc$_a$ & 96.04 & 96.13 & 96.39 & 96.48 \\
Acc$_t$ & 95.86 & 96.49 & 96.07 & 96.22\\
\midrule 
Sel$_a$ & 14.45 & 23.49 & 22.65 & 19.82 \\
Sel$_t$ & 31.68 & 31.82 & 37.31 & 38.51\\
\midrule
\multicolumn{5}{c}{SEM}  \\
\midrule
Neu$_t$ & 400/4\% & 400/4\% & 716/10\% & 307/10\% \\
Acc$_a$ & 92.09  & 92.64 & 91.94 & 93.29 \\
Acc$_t$ & 92.12 & 92.62 & 91.97 & 93.17 \\
\midrule 
Sel$_a$ & 5.77 & 14.03 & 12.78 & 11.18 \\
Sel$_t$ & 27.17 & 26.55 & 23.87 & 32.28\\
\midrule
\multicolumn{5}{c}{Chunking}  \\
\midrule
Neu$_t$ & 1000/10\% & 1000/10\% & 860/12\% & 983/32\% \\
Acc$_a$ & 95.01 & 94.62 & 93.43 & 93.14 \\
Acc$_t$ & 94.99 & 94.17 & 93.37 & 93.08 \\
\midrule 
Sel$_a$ & 16.30 & 22.77 & 24.42 & 18.13 \\
Sel$_t$ & 29.19 & 28.42 & 30.95 & 26.21\\
\midrule
\multicolumn{5}{c}{CCG}  \\
\midrule
Neu$_t$ & 1500/15\% & 1500/15\% & 2365/33\% & 1014/33\% \\
Acc$_a$ & 92.16  & 92.55 & 91.7 & 91.19 \\
Acc$_t$ & 92.36 & 92.39 & 91.39 & 90.95 \\
\midrule 
Sel$_a$ & 7.33 & 14.02 & 11.99 & 11.48 \\
Sel$_t$ & 15.06 & 24.15 & 18.32 & 17.88 \\
\bottomrule
\end{tabular}
}
\caption{Selecting minimal number of neurons for each downstream NLP task. Accuracy numbers reported on blind test-set (averaged over three runs) -- Neu$_a$ = Total number of neurons, Neu$_t$ = Top selected neurons, Acc$_a$ = Accuracy using all neurons, Acc$_t$ = Accuracy using selected neurons after retraining the classifier using selected neurons, Sel = Difference between linguistic task and control task accuracy when classifier is trained on all neurons (Sel$_a$) and top neurons (Sel$_t$).
}							
\label{tab:accuracy}		
\end{table}

\begin{table}[t]									
\centering		
\resizebox{\columnwidth}{!}{
\begin{tabular}{l|cccc}									
\toprule									
 & BERT & XLNet & T-ELMo & ELMo \\		
 \midrule
 Neu$_a$ & 9984  & 9984 & 7168 & 3072 \\
\midrule
\multicolumn{5}{c}{POS}  \\
\midrule
Neu$_t$ & 250/2.5\%  & 250/2.5\% & 215/3\% & 153/5\%\\
Acc$_a$ & 96.04 & 96.13 & 96.39 & 96.48 \\
Acc$_t$ & 93.70 & 95.72 & 94.92 & 94.45\\
\midrule
\multicolumn{5}{c}{SEM}  \\
\midrule
Neu$_t$ & 250/2.5\% & 400/4\% & 286/4\% & 307/5\% \\
Acc$_a$ & 92.09  & 92.64 & 91.94 & 93.29 \\
Acc$_t$ & 91.44 & 90.92 & 90.17 & 93.17 \\
\midrule
\multicolumn{5}{c}{Chunking}  \\
\midrule
Neu$_t$ & 600/6\% & 600/6\% & 430/6\% & 614/20\% \\
Acc$_a$ & 95.01 & 94.62 & 93.43 & 93.14 \\
Acc$_t$ & 93.53 & 92.83 & 92.28 & 91.79 \\
\midrule
\multicolumn{5}{c}{CCG}  \\
\midrule
Neu$_t$ & 698/7\% & 734/8\% & 716/10\% & 675/22\% \\
Acc$_a$ & 92.16  & 92.55 & 91.70 & 91.19 \\
Acc$_t$ & 91.73 & 91.11 & 89.79 & 89.08 \\
\bottomrule 
\end{tabular}
}
\caption{Selecting minimal number of neurons for each downstream NLP task with a looser threshold $\delta=2$. Accuracy numbers reported on blind test-set (averaged over three runs) -- Neu$_a$ = Total number of neurons, Neu$_t$ = Top selected neurons, Acc$_a$ = Accuracy using all neurons, Acc$_t$ = Accuracy using selected neurons after retraining the classifier using selected neurons.
}							
\label{tab:accuracy-lose}		
\end{table}


\subsection{Minimal Neuron Set}

Now that we have established correctness of the rankings, we apply the algorithm incrementally to select minimal neurons 
for each linguistic task that obtain a similar accuracy (we use a threshold $\delta = 0.5$)  as using the entire network (all the features). 
Identifying a minimal set of top neurons 
enables us to highlight: i) parts of the learned network where different linguistic phenomena are predominantly captured, ii) how localized or distributed information is with respect to different properties. 

Table~\ref{tab:accuracy} summarizes the results. Firstly we show that in all the tasks, selecting a subset of top N\% neurons and retraining the classifier can obtain a similar (sometimes even better) accuracy as using all the neurons (Acc$_a$) for classification as static features. For lexical tasks such as \textbf{POS} or \textbf{SEM} tagging, a very small number of neurons (roughly 400 i.e 4\% of features in BERT and XLNet) 
was found to be sufficient for achieving an accuracy (Acc$_t$) similar to oracle (Acc$_a$). More complex syntactic tasks such as \textbf{Chunking} and \textbf{CCG} tagging required larger sets of neurons (up to 2365 -- one third of the network in T-ELMo) to accomplish the same. 
It is interesting to see that all 
the models,  irrespective of 
their size, required a comparable number of selected neurons, in most of the cases.  On the \textbf{POS} and \textbf{SEM} tagging tasks, besides T-ELMo all other models use roughly the same number of neurons. T-ELMo required more neurons in \textbf{SEM} tagging to achieve the task. This could 
imply that 
knowledge of lexical semantics in T-ELMo  is distributed in more neurons. 
In an overall trend, ELMo generally needed fewer neurons 
while T-ELMo required more neurons 
compared to the other models to achieve oracle performance. Both 
these models are much smaller than BERT and XLNet. We did not observe any correlation, comparing results with the size of the models.



\paragraph{Control Tasks:} 
We use \emph{Selectivity} to further demonstrate that our probes (trained using the entire representation and selected neurons) do not memorize from word types but 
learned the underlying linguistic task. Recall that an effective probe is recommended to achieve high linguistic task accuracy and low control task accuracy. 
The results (see Table \ref{tab:accuracy}) show that selectivity with top neurons ($Sel_t$) is much higher than selectivity with all neurons $Sel_a$. It is evident that using all the neurons may contribute to memorization whereas higher selectivity with selected neurons indicates less memorization and efficacy of our neuron selection. 
%
We achieve high selectivity when selecting 400 neurons as in the case of POS and SEM. The chunking and CCG tasks require a lot more neurons with CCG requiring up to 33\% of the network. Here, the low selectivity indicates that while the information about CCG is distributed into several neurons, a 
set of random neurons may also be able to achieve a decent performance. 
%

\paragraph{Discussion:} 

Identifying neurons that are salient to a task 
has various potential applications such as task-specific model compression, by removing the irrelevant neurons with respect to the task or task-specific fine-tuning based on selected neurons. It is however tricky how to model this, for example one complexity is that zeroing out non-salient neurons in the lower layers directly affects any salient neurons in the subsequent layers. A rather direct application to our work is 
efficient feature-based transfer learning,
which has 
shown to be a viable alternative to the fine-tuning approach \citep{peters-etal-2019-tune}.   Feature-based  
approach uses contextualized embeddings learned from pre-trained models as static feature vectors in the down-stream  classification task.  Classifiers with large contextualized vectors are not only cumbersome to train, but also inefficient during inference. They have also been shown to be sub-optimal when supervised data is insufficient \cite{hameed}.  BERT-large, for example, is trained with 19,200 (25 layers $\times$ 768 dimensions) features. Reducing the feature set to a smaller number can lead to faster training of the classifier and efficient inference. Earlier (in Table \ref{tab:accuracy}) we obtained minimal set of neurons with a very tight threshold of $\delta=0.5$. By allowing a loser threshold, say $\delta=2$, we can reduce the set of minimal neurons to improve the efficiency even more. See Table \ref{tab:accuracy-lose} for results. For more on this, we refer interested readers to look at 
\newcite{dalvi-2020-CCFS}, where we explored this more formally, expanding our study to the sentence-labeling GLUE tasks \cite{wang-etal-2018-glue}.



\begin{figure*}[t]
    \centering
    \begin{subfigure}[b]{0.23\linewidth}
    \centering
    \includegraphics[width=\linewidth]{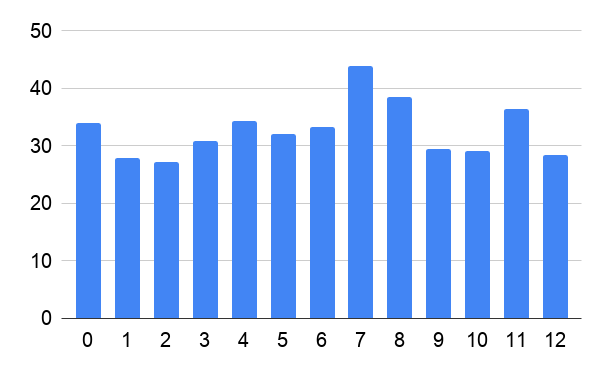}
    \caption{POS -- BERT}
    \label{fig:posbert}
    \end{subfigure}
    \begin{subfigure}[b]{0.23\linewidth}
    \centering
    \includegraphics[width=\linewidth]{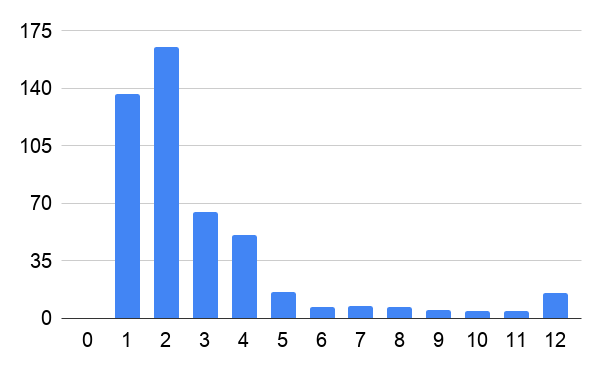}
    \caption{POS -- XLNet}
    \label{fig:posxlnet}
    \end{subfigure}    
    \begin{subfigure}[b]{0.23\linewidth}
    \centering
    \includegraphics[width=\linewidth]{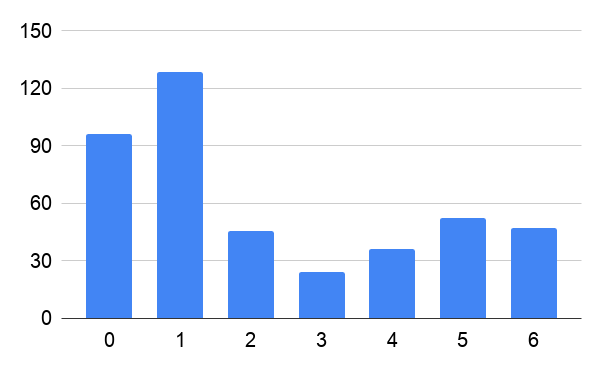}
    \caption{POS -- T-ELMo}
    \label{fig:poscalypso}
    \end{subfigure}
    \begin{subfigure}[b]{0.23\linewidth}
    \centering
    \includegraphics[width=\linewidth]{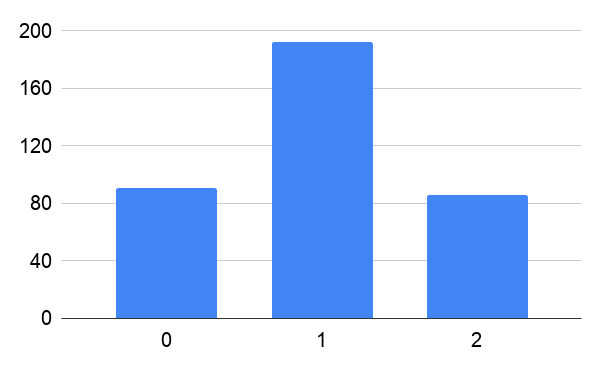}
    \caption{POS -- ELMo}
    \label{fig:poselmo}
    \end{subfigure}
    
    \begin{subfigure}[b]{0.23\linewidth}
    \centering
    \includegraphics[width=\linewidth]{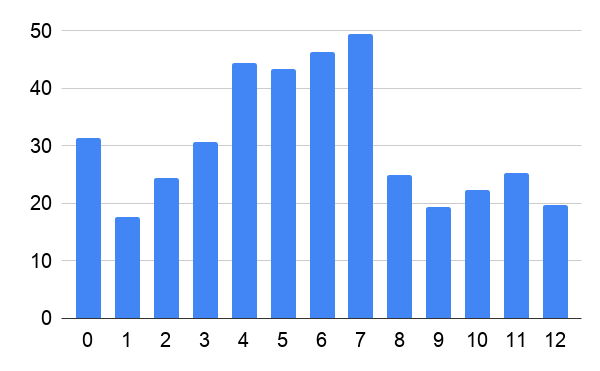}
    \caption{SEM -- BERT}
    \label{fig:sembert}
    \end{subfigure}
    \begin{subfigure}[b]{0.23\linewidth}
    \centering
    \includegraphics[width=\linewidth]{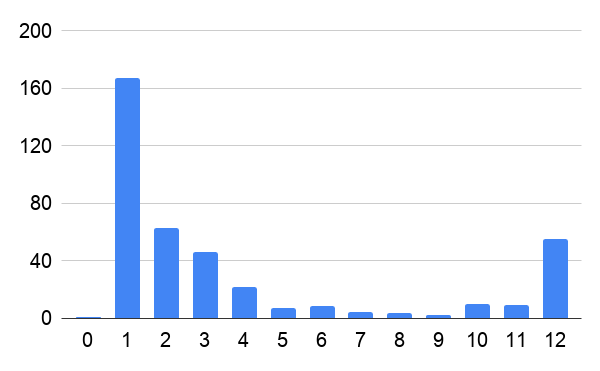}
    \caption{SEM -- XLNet}
    \label{fig:semxlnet}
    \end{subfigure}    
    \begin{subfigure}[b]{0.23\linewidth}
    \centering
    \includegraphics[width=\linewidth]{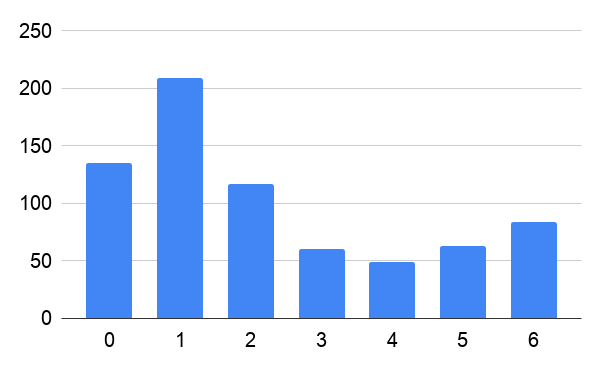}
    \caption{SEM -- T-ELMo}
    \label{fig:semcalypso}
    \end{subfigure}
    \begin{subfigure}[b]{0.23\linewidth}
    \centering
    \includegraphics[width=\linewidth]{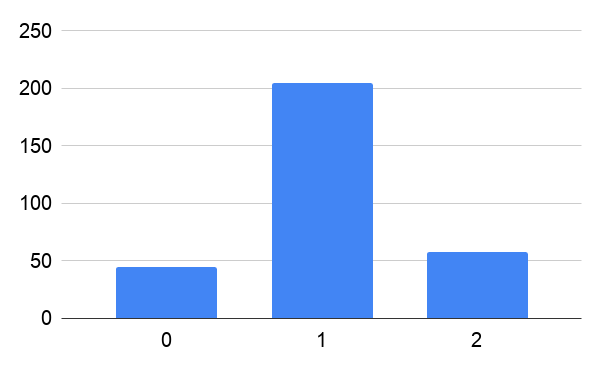}
    \caption{SEM -- ELMo}
    \label{fig:semelmo}
    \end{subfigure}    
    

    \begin{subfigure}[b]{0.23\linewidth}
    \centering
    \includegraphics[width=\linewidth]{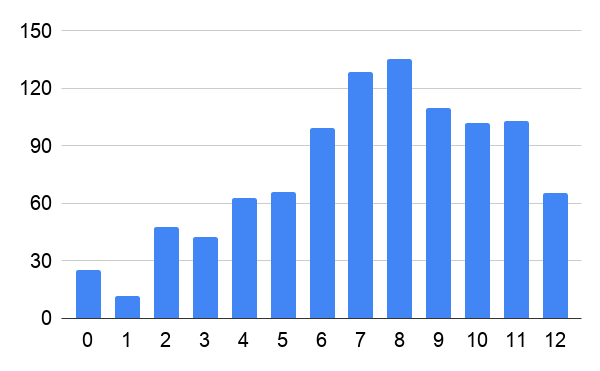}
    \caption{Chunking -- BERT}
    \label{fig:chunkingbert}
    \end{subfigure}
    \begin{subfigure}[b]{0.23\linewidth}
    \centering
    \includegraphics[width=\linewidth]{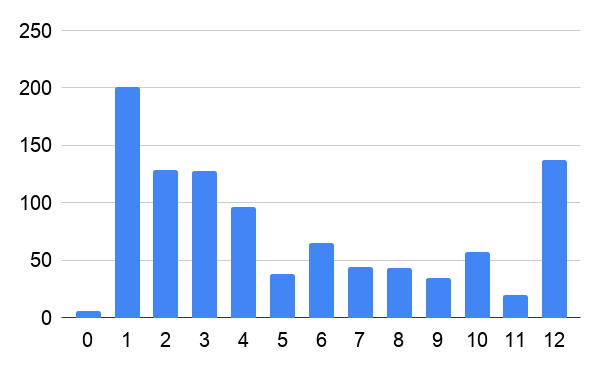}
    \caption{Chunking -- XLNet}
    \label{fig:chunkingxlnet}
    \end{subfigure}    
    \begin{subfigure}[b]{0.23\linewidth}
    \centering
    \includegraphics[width=\linewidth]{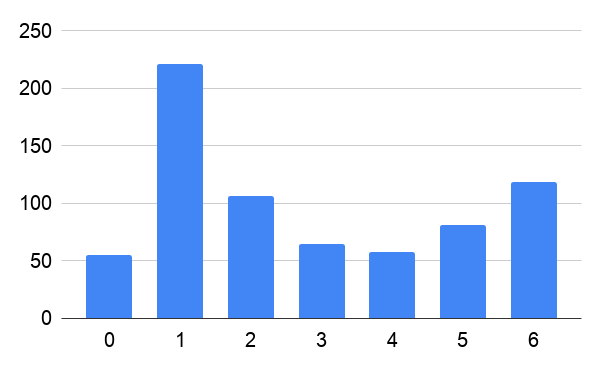}
    \caption{Chunking -- T-ELMo}
    \label{fig:chunkingcalypso}
    \end{subfigure}
    \begin{subfigure}[b]{0.23\linewidth}
    \centering
    \includegraphics[width=\linewidth]{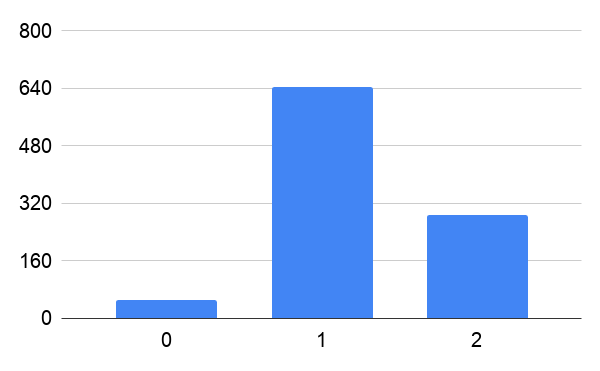}
    \caption{Chunking -- ELMo}
    \label{fig:chunkingelmo}
    \end{subfigure}    

    \begin{subfigure}[b]{0.23\linewidth}
    \centering
    \includegraphics[width=\linewidth]{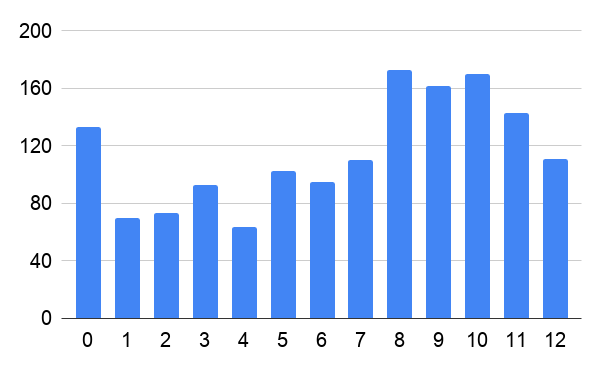}
    \caption{CCG -- BERT}
    \label{fig:ccgbert}
    \end{subfigure}
    \begin{subfigure}[b]{0.23\linewidth}
    \centering
    \includegraphics[width=\linewidth]{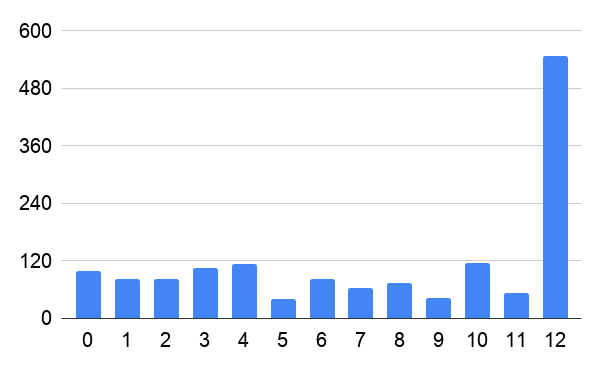}
    \caption{CCG -- XLNet}
    \label{fig:ccgxlnet}
    \end{subfigure}    
    \begin{subfigure}[b]{0.23\linewidth}
    \centering
    \includegraphics[width=\linewidth]{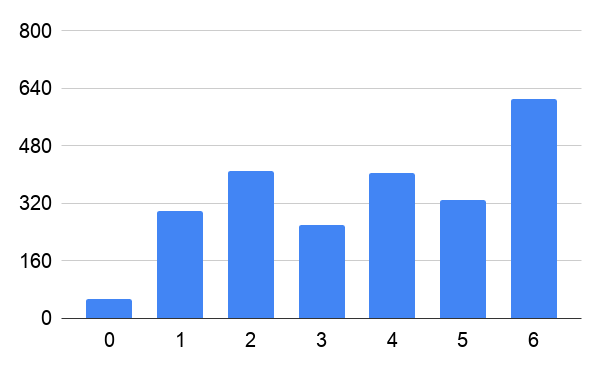}
    \caption{CCG -- T-ELMo}
    \label{fig:ccgcalypso}
    \end{subfigure}
    \begin{subfigure}[b]{0.23\linewidth}
    \centering
    \includegraphics[width=\linewidth]{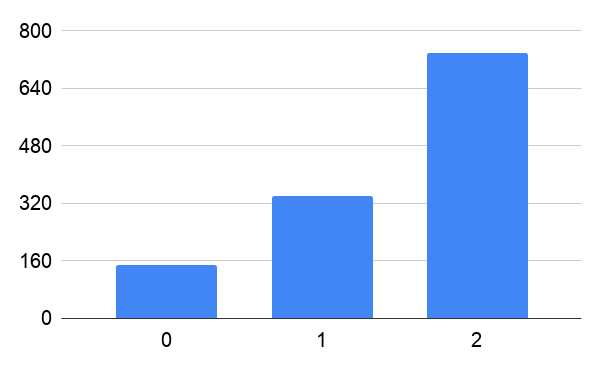}
    \caption{CCG -- ELMo}
    \label{fig:ccgelmo}
    \end{subfigure}
    
    \caption{How top neurons spread across different layers for each task? X-axis = Layer number, Y-axis = Number of neurons selected from that layer}
    \label{fig:layerwise}
\end{figure*}

\section{Analysis}
\label{sec: lw-analysis}
\subsection{Layer-wise Distribution}
\label{sec: lwDistro}

Previous work on analyzing deep neural networks 
analyzed how individual layers contribute towards a downstream task \cite{liu-etal-2019-linguistic,kim2020pretrained,belinkov-etal-2020-analysis}. 
Here we observe how the neurons, selected from the entire network, spread across different layers of the model. Such an analysis gives an alternative view of which layers contribute predominantly towards different tasks.
Figure \ref{fig:layerwise} presents the results.
In most cases, lexical tasks such as learning morphology (POS tagging) and word semantics (SEM tagging) are dominantly captured by the neurons at lower layers, whereas the more complicated task of modeling syntax (CCG supertagging) is taken care of at the final layer.  
An exception to this overall pattern is the BERT model.
Top neurons in BERT spread across all the layers, unlike other models where top neurons (for a particular task) are contributed by fewer layers.
This reflects that every layer in BERT possesses neurons that specialize in learning particular language properties, while other models have 
designated layers that specialize in learning those language properties. 
Different from other models, neurons in the embedding layer show minimum contribution in XLNet consistently across the tasks. 
Let us analyze the 
results with respect to each linguistic task. 

\paragraph{POS Tagging:} 
Every layer in BERT and ELMo contributed towards the top neurons, while the distribution is dominated by lower layers in XLNet and T-ELMo, with an exception of XLNet not choosing any neurons from the embedding layer. 

\paragraph{SEM Tagging:} Similar to POS, all layers of BERT contributed to the list of top neurons. However, the middle layers showed the most contribution (see layer numbers 4--7 in Figure \ref{fig:sembert}). This is in line with 
\newcite{liu-etal-2019-linguistic} 
who found middle and higher middle layers to give optimal results for the semantic tagging task.  On XLNet, T-ELMo and ELMo, the first layer after the embedding layer 
got the largest share of the top neurons of SEM. This trend is consistent across other tasks,  i.e., the core linguistic information is learned earlier in the network with an exception of BERT, which distributes information across the network.



\paragraph{Chunking Tagging:} The overall 
pattern remained similar in the task of chunking. 
Notice however, a shift in pattern -- the contribution from lower layers decreased compared to previous tasks, 
in the case of BERT. For example, in the SEM task, top neurons were dominantly contributed from lower and middle layers, in chunking middle and higher layers contributed most. This could be attributed to the fact that chunking is a more complex syntactic task and is learned at relatively higher layers.

\paragraph{CCG Supertagging:} Compared to chunking, CCG supertagging is a richer syntactic tagging task, almost equivalent to parsing \cite{J99-2004}. 
The complexity of the task is evident in our results as there is a clear shift in the distribution of top neurons moving from middle to higher layers. 
The only exception again is the BERT model where this information is 
well spread across the network, but still dominantly preserved in the final layers.

\paragraph{Discussion:} Our results are in line with and reinforce the layer-wise analysis presented in \newcite{liu-etal-2019-linguistic}. 
However, unlike their work and all other work on layer-wise probing analysis, which trains a classifier on each layer individually to compare the results, our method trains a single classifier on all layers concatenated 
to analyze which layers contribute most to the task based on the most relevant selected features. This makes the playing field even and results in a sharper analysis. For example, \newcite{liu-etal-2019-linguistic} showed layer 1 in Transformer-ELMo to give the best result on the task of predicting POS tags; however, layers 2 and 3 almost give similar accuracy (see Appendix D1 in their paper). Based on these results, one cannot confidently claim that the task of POS is predominantly captured at layer 1. However, our method clearly shows this result (see Figure \ref{fig:poscalypso}).

\subsection{Localization versus Distributedness}
\label{sec:locVsDistr}


\begin{figure}[t]
	\centering
	\includegraphics[width=\linewidth]{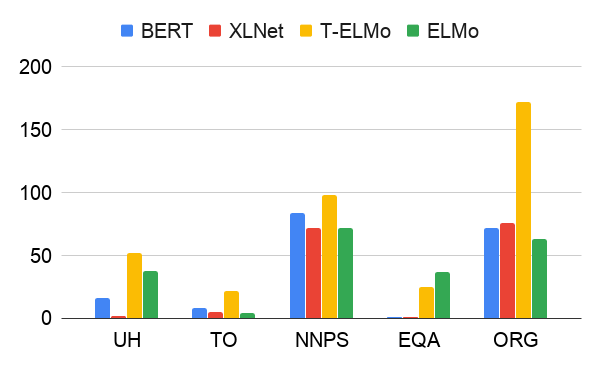} 
	\caption{Number of neurons per label: Some properties (e.g., interjections) are localized in fewer neurons, while others (e.g., nouns) are more distributed.   
	Y-axis = number of neurons per label}
  \vspace{-4mm}
	\label{fig:neurons_per_architecture}
\end{figure}

\begin{figure*}[t]
    \center
    \begin{subfigure}[b]{0.49\linewidth}
    \centering
    \includegraphics[width=\linewidth]{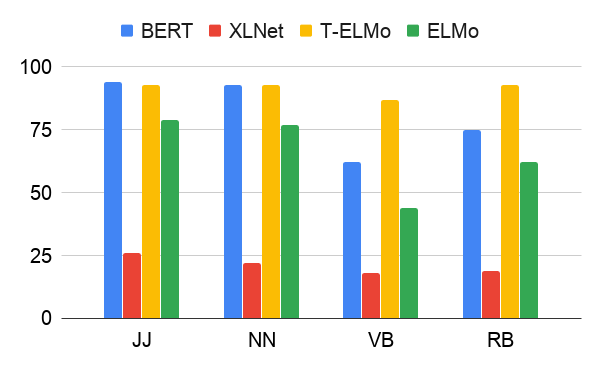}
    \caption{POS Tagging}
    \label{fig:pos_total_neurons}
    \end{subfigure}
    \begin{subfigure}[b]{0.49\linewidth}
    \centering
    \includegraphics[width=\linewidth]{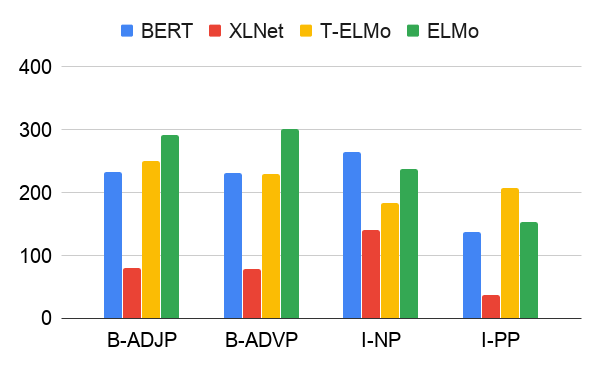}
    \caption{Chunking Tagging}
    \label{fig:sem_total_neurons}
    \end{subfigure}    
    \caption{Top neurons in XLNet are more localized towards individual properties compared to other architectures}
    \label{fig:neurons_per_task}
\end{figure*}


Next we study how localized or distributed different properties  are within a linguistic task  (for example nouns or verbs in POS tagging, location in semantic tagging), and across different architectures. Remember that the ranking algorithm extracts neurons for each label $t$ (e.g. LOC:location or EVE:event categories in semantic tagging) in task $T$, sorted based on absolute weights. The final rankings are obtained by selecting from each label using the neuron ranking algorithm as described in Section \ref{sec:methodology}. This allows us to analyze how localized or distributed a property is, based on the number of neurons that are selected for  each label in the task.

\paragraph{Property-wise:} We found that while many properties are distributed, i.e., a large group of neurons is used to predict a label, some properties such as functional or unambiguous words 
that do not require contextual information are learned using fewer neurons. For example, \textbf{UH} (interjections) or the \textbf{TO} particle required fewer neurons across architectures compared to \textbf{NNPS} (proper noun; plural) in the task of POS tagging (Figure \ref{fig:neurons_per_architecture}).   Similarly \textbf{EQA} (equating property, e.g., \textit{\textbf{as} tall as you}) is handled with fewer neurons compared to \textbf{ORG} (organization property). We observed a similar behavior in the task of chunking, with \textbf{I-PRT} (particles inside of a chunk) requiring fewer neurons across different architectures. On the contrary, \textbf{B-VP} (beginning of verb phrase) required plenty many. 

\paragraph{Layer-wise:} Previously we analyzed each linguistic task in totality. We now study whether individual properties (e.g., adjectives) are localized or well distributed across layers in different architectures. We observed interesting cross architectural similarities, for example the neurons that predict the foreign words (\textbf{FW}) property were predominantly localized in final layers (BERT: 13, XLNET: 11, T-ELMo: 7, ELMo:3) of the network in all the understudied architectures. In comparison, the neurons that capture common class words such as adjectives (\textbf{JJ}) and locations (\textbf{LOC}) are localized in lower layers (BERT: 0, XLNET: 1, T-ELMo: 0, ELMo:1). 
In some cases, we did find variance, for example personal pronouns (\textbf{PRP}) in POS tagging and event class (\textbf{EXC}) in semantic tagging were handled at different layers across different architectures. See Appendix \ref{subsec:xlnet} for all labels.

\paragraph{Architecture-wise:} 
We found that top neurons in XLNet are more localized towards individual properties compared to other architectures where top neurons are shared across multiple properties. We demonstrate this in Figure 
\ref{fig:neurons_per_task}. Notice how the number of neurons for different labels\footnote{Figure \ref{fig:neurons_per_task} only displays selected properties, but the pattern holds across all properties. See  Appendix \ref{subsec:xlnet}.} is much smaller in the case of XLNet, although roughly the same number of total neurons (400 for POS tagging and 960 for chunking on average; see Table \ref{tab:accuracy}) were required by all pre-trained models to carry out a task. 
This means that in XLNet neurons  are 
exclusive towards specific properties 
compared to other architectures where neurons are 
shared between multiple properties. Such a trait in XLNet can be potentially helpful in 
predicting the behavior of the system as it is easier to isolate neurons that are designated toward specific phenomena.

\section{Related Work}

Rise of neural network has seen a subsequent rise of interpretability of these models. 
Researchers have explored visualization methods to analyze learned representations \cite{karpathy2015visualizing,kadar2016representation}, attention heads \cite{clark-etal-2019-bert, vig-2019-multiscale} of language compositionality \cite{li-etal-2016-visualizing} etc. While such visualizations illuminate the inner workings of the network, they are often qualitative in nature and somewhat anecdotal. 

A more commonly used approach tries to 
provide a quantitative analysis by correlating parts of the neural network with linguistic properties, for example by training a classifier to predict a feature of interest~\cite{adi2016fine,conneau2018you}. Please refer to \newcite{belinkov-glass-2019-analysis} for a comprehensive survey of work done in this direction.
\newcite{liu-etal-2019-linguistic} used probing classifiers for investigating the contextualized representations learned from a variety of neural language models on numerous word level linguistic tasks.  
A similar analysis was carried by \newcite{tenney2018what} on a variety of sub-sentence linguistic tasks. We extend this line of work to carry out a more fine-grained neuron level analysis of neural language models.

Our work is most similar to \newcite{dalvi:2019:AAAI} who conducted neuron analysis of representations learned from sequence-to-sequence machine translation models. Our work is different from them in that i) we carry out analysis on a wide range of architectures which are deeper and more complicated than RNN-based models and illuminate interesting insights, ii) we automated the grid-search criteria to select the regularization parameters, compared to manual selection of lambdas, which is cumbersome and error-prone. In contemporaneous work, \newcite{suau2020finding} used \emph{max-pooling} to identify relevant neurons (aka \emph{Expert units}) in pre-trained models, with respect to a specific concept (for example word-sense). 

A pitfall to the approach of probing classifiers is whether the probe is faithfully reflecting the property of the representation or just learned the task? \newcite{hewitt-liang-2019-designing} defined control tasks to analyze the role of training data and lexical memorization in probing experiments. \newcite{voita2020informationtheoretic} proposed an alternative that measures \emph{Minimal Description Length} of labels given representations. It would be interesting to see how a probe's complexity in their work (code length) compares with the number of selected neurons according to our method. The results are consistent at least in the ELMo POS example, where layer 1 was shown to have the shortest code length in their work. In our case, most top neurons are selected from layer 1 (see Figure \ref{fig:poselmo} for example).
\newcite{pimentel2020informationtheoretic} discussed the complexity of the probes and argued for using highest performing probes for tighter estimates. However, complex probes are difficult to analyze. Linear models are preferable due to their explainability; especially in our work, as we use the learned weights as a proxy to get a measure of the importance of each neuron. We used linear classifiers with control tasks as described in \newcite{hewitt-liang-2019-designing}. Although we mainly used probing accuracy to drive the neuron selection in this work, and \emph{Selectivity} only to demonstrate that our results reflect the property learned by representations and not probe's  capacity to learn -- an interesting idea would be to use selectivity itself to drive the investigation. However, it is not trivial how to optimize for selectivity as it cannot be controlled/tuned directly -- for example, removing some neurons may decrease accuracy but may not change selectivity. We leave this exploration for future work.  

Probing classifiers require supervision for the linguistic tasks of interest with annotations, limiting their applicability.  \newcite{bau2018identifying} used unsupervised approach to identify
salient neurons in neural machine translation 
and manipulated translation output by controlling these neurons. Recently, \newcite{wu:2020:acl} measured similarity of  internal representations and attention across prominent contextualized representations (from BERT, ELMo, etc.). They found that different architectures 
have similar representations, but different individual neurons.

\section{Conclusion}
\label{sec:conclusion}

We analyzed individual neurons 
across a variety of neural language models 
using linguistic correlation analysis 
on the task of predicting core linguistic properties (morphology, syntax and semantics). Our results reinforce previous findings and also illuminate further insights: i) while the information in neural language models is massively distributed, it is possible to extract a small number of features to carry out a downstream NLP task, ii) the number of extracted features varies based on the complexity of the task, iii) the neurons that learn word morphology and lexical semantics are predominantly found in the lower layers of the network, whereas the ones that learn syntax are at the higher layers, with the exception of BERT, where neurons were spread across the entire network, iv) closed-class words (for example interjections) are handled using fewer neurons compared to polysemous words (such as nouns and adjectives), v) features in XLNet are more localized towards individual properties as opposed to other architectures where neurons are distributed across many properties. A direct application of our analysis 
is efficient feature-based transfer learning from large-scale neural language models: i) identifying that most relevant features for a task are contained in layer $x$ reduces the forward-pass to that layer, ii) reducing the feature set decreases the time to train a classifier and also its inference. 
We refer interested readers to see our work presented in \newcite{dalvi-2020-CCFS} for more details.

\section*{Acknowledgements}

We thank the anonymous reviewers for their feedback on the earlier draft of this paper. 
This research was carried out in collaboration between the Qatar Computing Research Institute (QCRI) and the 
MIT Computer Science and Artificial Intelligence Laboratory (CSAIL). Y.B.\ was also supported by the Harvard Mind, Brain, and Behavior Initiative (MBB).

\bibliography{acl2020}
\bibliographystyle{acl_natbib}

\newpage
\clearpage
\appendix

\section{Appendices}
\subsection{Data and Representations}
\label{subsec:data}

We used standard splits for training, development and test data for the 4 linguistic tasks (POS, SEM, Chunking and CCG super tagging) that we used to carry out our analysis on. The splits to preprocess the data are available through git repository\footnote{\url{https://github.com/nelson-liu/contextual-repr-analysis}} released with \newcite{liu-etal-2019-linguistic}. See Table \ref{tab:dataStats} for statistics. We obtained the understudied pre-trained models from the authors of the paper, through personal communication.

\begin{table}[ht]									
\centering					
\footnotesize
\resizebox{\columnwidth}{!}{									
    \begin{tabular}{l|cccc}									
    \toprule									
Task    & Train & Dev & Test & Tags \\		
\midrule
    POS & 36557 & 1802 & 1963 & 44 \\
    SEM & 36928 & 5301 & 10600 & 73 \\
    Chunking &  8881 &  1843 &  2011 & 22 \\
    CCG &  39101 & 1908 & 2404 & 1272 \\
    \bottomrule
    \end{tabular}
    }
    \caption{Data statistics (number of sentences) on training, development and test sets using in the experiments and the number of tags to be predicted}

\label{tab:dataStats}						
\end{table}

\subsection{Hyperparameters}
\label{subsec:hyperparameters}

We use elastic-net based regularization to control the trade-off between 
selecting focused individual neurons versus group of neurons while maintaining the original accuracy of the classifier without any regularization. We do a grid search on $L_1$ and $L_2$ ranging from values $0 \dots 1e^{-7}$. See Table \ref{tab:lambdas} for the optimal values for each task across different architectures.

\begin{table}[ht]									
\centering					
\resizebox{\columnwidth}{!}{									
    \begin{tabular}{l|cccc}									
    \toprule									
    & BERT & XLNet & T-ELMo & ELMo \\		
    \midrule
    \multicolumn{5}{c}{L1 , L2   = $\lambda_1$, $\lambda_2$ }  \\
    \toprule
    POS & .001, .01 & .001, .01 & .001, .001 & .001, .0001 \\
    SEM & .001, .01 & .001, .01 & .001, .001 & .001, .0001 \\
    Chunk & $1e^{-4}, 1e^{-5}$  & $1e^{-4}, 1e^{-4}$ & .001, .001 & .001, .01 \\
    CCG & $1e^{-5}, 1e^{-6}$ & $1e^{-5}, 1e^{-6}$ & $1e^{-4}, 1e^{-6}$ & $1e^{-5}, 1e^{-6}$ \\
    \bottomrule
    \end{tabular}
    }
    \caption{Best elastic-net lambdas parameters for each task}

\label{tab:lambdas}						
\end{table}

\subsection{Infrastructure and Run Time}
\label{subsec:runTime}

Our experiments were run on NVidia GeForce GTX TITAN X GPU card. Grid search for finding optimal lambdas is expensive when optimal number of neurons for the task are unknown. Running grid search would take $\mathcal{O} (M N^2)$ where $M = 100 $ (if we try increasing number of neurons in each step by 1\%) and $N = 0, 0.1, \dots 1e^{-7}$. We fix the $M=20\%$ to find the best regularization parameters first reducing the grid search time to $\mathcal{O} (N^2)$ and find the optimal number of neurons in a subsequent step with $\mathcal{O} (M)$. The overall running time of our algorithm therefore is $\mathcal{O} (M + N^2)$. This varies a lot in terms of wall-clock computation, based on number of examples in the training data, number of tags to be predicted in the downstream task. Including a full forward pass over the pre-trained model to extract the contextualized vector, and running the grid search algorithm to find the best hyperparameters and minimal set of neurons took on average 12 hours ranging from 3 hours (for POS with ELMo experiment) to 18 hours (for CCG with BERT).

\subsection{Ablation Study}
\label{subsec:ablationStudy}

We reported accuracy numbers on ablating top, random and bottom neurons in the trained classifier, on blind test-set in the main body. In Table \ref{tab:classifier_ablation_mask_out_dev}, we report results on development tests.

\begin{table}[t]									
\centering					
\footnotesize
\resizebox{\columnwidth}{!}{									
    \begin{tabular}{l|cccc}									
    \toprule									
    & BERT & XLNet & T-ELMo & ELMo \\		
    \midrule
    \multicolumn{5}{c}{POS}  \\
    \toprule
    All & 96.10 & 96.38 & 96.61 & 96.45 \\
    Top & 90.32 & 93.07 & 92.13 & 85.03 \\
    Rand & 29.43 & 57.32 & 49.14 & 32.18 \\
    Bot & 17.99 & 45.61 & 23.01 & 17.36 \\
    \midrule 
    \multicolumn{5}{c}{SEM}  \\
    \midrule
    All & 92.63  & 92.16 & 92.40 & 93.35 \\
    Top & 85.17 & 90.91 & 84.13 &  83.01\\
    Rand & 65.12 & 71.11 & 65.11 & 74.18\\
    Bot & 58.19 & 26.11 & 35.99 & 57.11 \\
    \midrule 
    \multicolumn{5}{c}{Chunking}  \\
    \midrule
    All & 95.11 & 94.19 & 93.93 & 93.85 \\
    Top & 90.13 & 90.03 & 88.13 & 83.12 \\
    Rand & 74.12 & 75.63 & 78.19 & 71.48 \\
    Bot & 64.13 & 45.43 & 47.16 & 65.12 \\
    \midrule 
    \multicolumn{5}{c}{CCG}  \\
    \toprule
    All & 92.23  & 92.43 & 91.66 & 91.23 \\
    Top & 75.61 & 76.31 & 71.22 & 68.09 \\
    Rand & 70.01 & 63.11 & 68.03 & 41.37\\
    Bot & 61.12 & 62.31 & 67.99 & 30.12  \\
    \bottomrule
    \end{tabular}
    }
    \caption{Ablation Study: Selecting all, top, random (rand) and bottom (bot) 20\% neurons and zeroing-out remaining to evaluate classifier accuracy on dev test (averaged over three runs).}

\label{tab:classifier_ablation_mask_out_dev}						
\end{table}

\subsection{Minimal Neuron Set}
\label{subsec:minimalSet}

We reported minimal number of neurons required to obtain oracle accuracy in the main body, along with the results on \emph{Selectivity}. In Table \ref{tab:accuracy_dev}, we report results on development tests.

\begin{table}[t]									
\centering		
\resizebox{\columnwidth}{!}{
\begin{tabular}{l|cccc}									
\toprule									
 & BERT & XLNet & T-ELMo & ELMo \\		
 \midrule
 Neu$_a$ & 9984  & 9984 & 7168 & 3072 \\
\midrule
\multicolumn{5}{c}{POS}  \\
\midrule
Neu$_t$ & 400/4\%  & 400/4\% & 430/6\% & 368/12\%\\
Acc$_a$ & 96.10 & 96.38 & 96.61 & 96.45 \\
Acc$_t$ & 96.48 & 96.52 & 96.33 & 96.07\\
\midrule 
Sel$_a$ & 15.51 & 23.43 & 22.69 & 19.12 \\
Sel$_t$ & 31.81 & 31.62 & 37.61 & 38.52\\
\midrule
\multicolumn{5}{c}{SEM}  \\
\midrule
Neu$_t$ & 400/4\% & 400/4\% & 716/10\% & 307/10\% \\
Acc$_a$ & 92.63  & 92.16 & 92.40 & 93.35  \\
Acc$_t$ & 92.19 & 92.59 & 92.17 & 93.21 \\
\midrule 
Sel$_a$ & 5.82 & 14.01 & 12.19 & 11.37 \\
Sel$_t$ & 27.19 & 26.46 & 23.97 & 32.33\\
\midrule
\multicolumn{5}{c}{Chunking}  \\
\midrule
Neu$_t$ & 1000/10\% & 1000/10\% & 860/12\% & 983/32\% \\
Acc$_a$ & 95.11 & 94.19 & 93.93 & 93.85 \\
Acc$_t$ & 95.07 & 94.13 & 93.61 & 93.48 \\
\midrule 
Sel$_a$ & 16.33 & 22.87 & 24.31 & 18.09 \\
Sel$_t$ & 29.32 & 28.19 & 31.05 & 26.38\\
\midrule
\multicolumn{5}{c}{CCG}  \\
\midrule
Neu$_t$ & 1500/15\% & 1500/15\% & 2365/33\% & 1014/33\% \\
Acc$_a$ & 92.23  & 92.43 & 91.66 & 91.23 \\
Acc$_t$ & 92.13 & 92.49 & 91.89 & 91.09 \\
\midrule 
Sel$_a$ & 7.48 & 14.21 & 11.42 & 11.99 \\
Sel$_t$ & 15.91 & 24.82 & 18.31 & 17.34 \\
\bottomrule
\end{tabular}
}
\caption{Selecting minimal number of neurons for each downstream NLP task. Accuracy numbers reported on dev test (averaged over three runs) -- Neu$_a$ = Total number of neurons, Neu$_t$ = Top selected neurons, Acc$_a$ = Accuracy using all neurons, All$_t$ = Accuracy using selected neurons after retraining the classifier using selected neurons, Sel = Difference between linguistic task and control task accuracy when classifier is trained on all neurons (Sel$_a$) and top neurons (Sel$_t$).
}							
\label{tab:accuracy_dev}						
\end{table}

\subsection{Localized versus Distributed Labels}
\label{subsec:localized}

In Section \ref{sec: lwDistro} we only showed number of features learned for selected labels in each task. Figure \ref{fig:full_neurons_per_task} shows results for all the tags across different tasks. The results show that some tags are localized and captured by a focused set of neurons while others are distributed and learned within a large set of neurons.

\subsection{XLNet versus Others}
\label{subsec:xlnet}

\begin{figure*}[t]
    \center
    \begin{subfigure}[b]{0.75\linewidth}
    \centering
    \includegraphics[width=\linewidth]{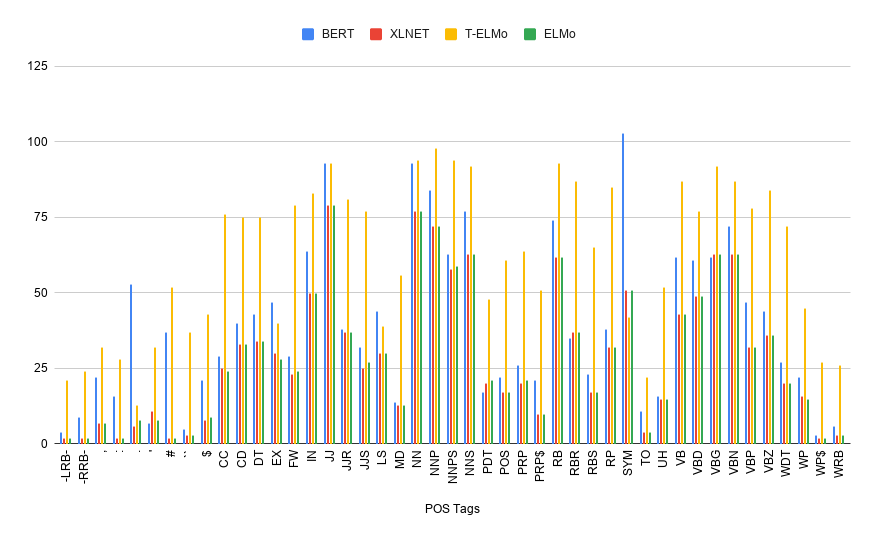}
    \caption{POS Tagging}
    \label{fig:pos_full_neurons}
    \end{subfigure}
    \begin{subfigure}[b]{0.75\linewidth}
    \centering
    \includegraphics[width=\linewidth]{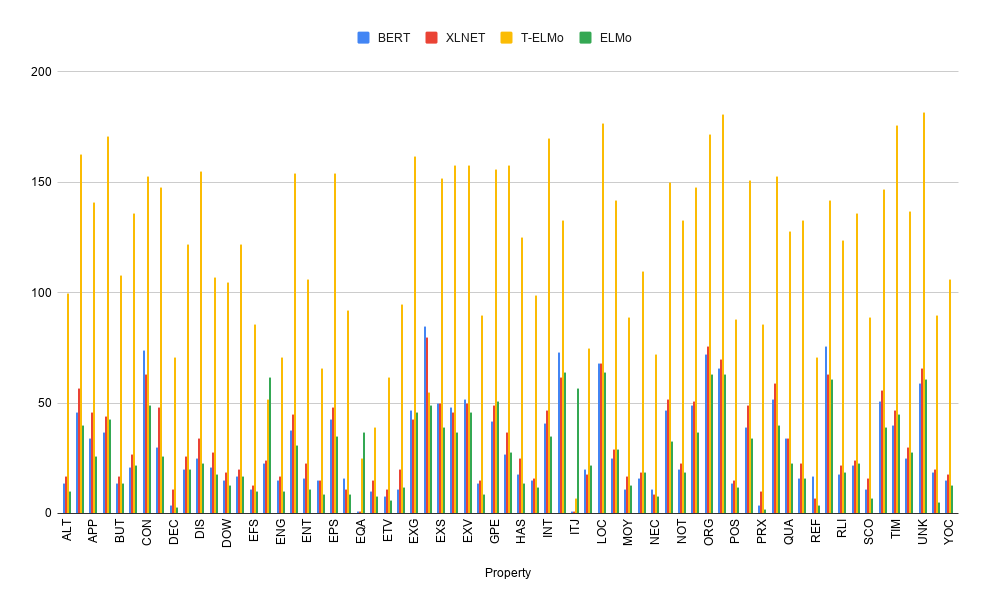}
    \caption{SEM Tagging}
    \label{fig:sem_full_neurons}
    \end{subfigure}
    \begin{subfigure}[b]{0.75\linewidth}
    \centering
    \includegraphics[width=\linewidth]{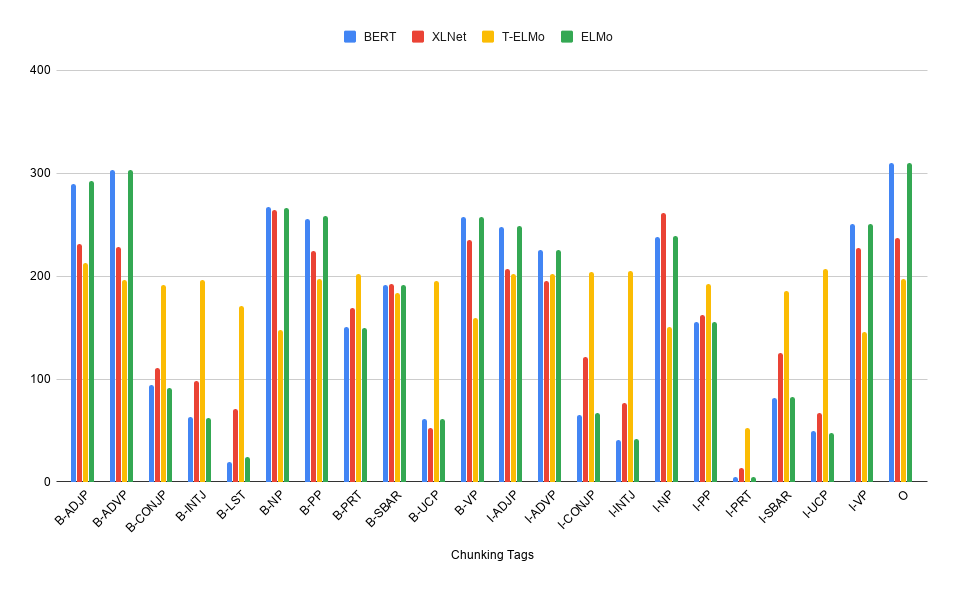}
    \caption{Chunking Tagging}
    \label{fig:sem_full_neurons}
    \end{subfigure}
    \caption{Number of neurons per label across architectures}
    \label{fig:full_neurons_per_task}
\end{figure*}

Notice in Figure \ref{fig:full_neurons_per_task} that neurons required by each label in XLNet (red bars) are strikingly small compared to other architectures specifically T-ELMo (yellow bars). This is interesting given the fact that total number of neurons required by some of the tasks are very similar. For example task of POS tagging required 400 neurons for BERT and XLNet, 320 for ELMo and 430 in T-ELMo. This means that neurons in XLNet are mutually exclusive towards the properties whereas in other architectures neurons are shared across multiple properties. Due to large tag set (1272 tags) in CCG super tagging, it is not possible to include it among figures.

\subsection{Layer-wise Distribution}
\label{subsec:layerWiseDistro}

\begin{figure*}[t]
    \center
    \begin{subfigure}[b]{0.75\linewidth}
    \centering
    \includegraphics[width=\linewidth]{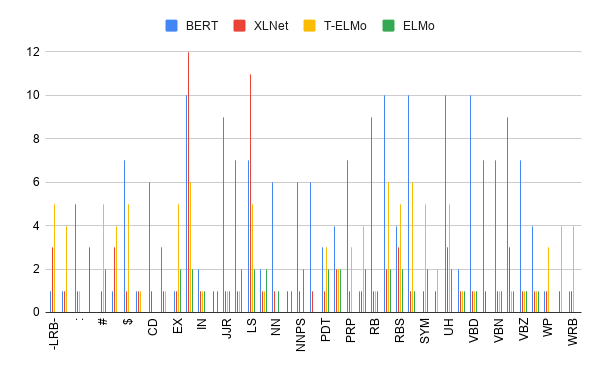}
    \caption{POS Tagging}
    \label{fig:pos_whichLayer}
    \end{subfigure}
    \begin{subfigure}[b]{0.75\linewidth}
    \centering
    \includegraphics[width=\linewidth]{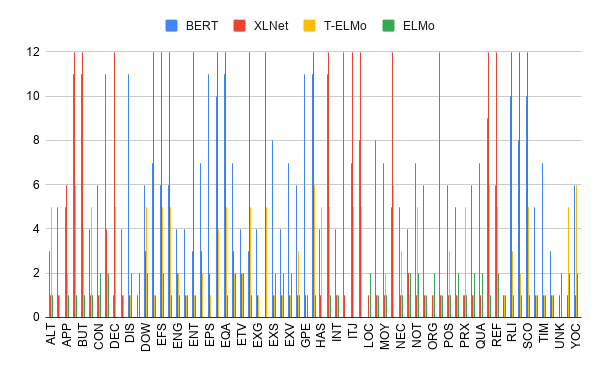}
    \caption{SEM Tagging}
    \label{fig:sem_whichLayer}
    \end{subfigure}
    \begin{subfigure}[b]{0.75\linewidth}
    \centering
    \includegraphics[width=\linewidth]{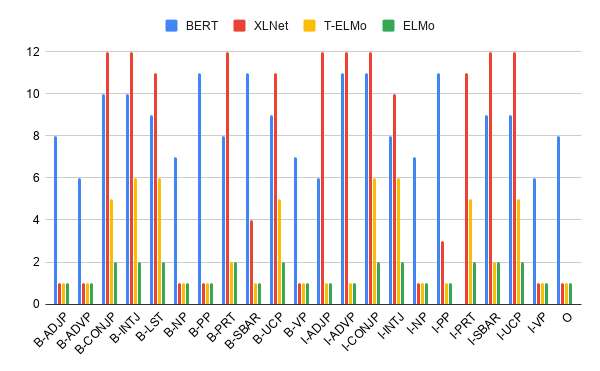}
    \caption{Chunking Tagging}
    \label{fig:chunking_whichLayer}
    \end{subfigure}
    \caption{Layer that predominately captures each label}
    \label{fig:whichLayer}
\end{figure*}

In Section \ref{sec:locVsDistr} we showed labels are captured dominantly at which layers for a few labels. In Figure \ref{fig:chunking_whichLayer} we show all labels and which layers they are predominantly captured at, across different architectures.

\end{document}

%% file: math_commands.tex
\usepackage{amsmath,amsfonts,bm}

\def\eqref#1{equation~\ref{#1}}

\def\1{\bm{1}}

\DeclareMathAlphabet{\mathsfit}{\encodingdefault}{\sfdefault}{m}{sl}
\SetMathAlphabet{\mathsfit}{bold}{\encodingdefault}{\sfdefault}{bx}{n}

\def\sD{{\mathbb{D}}}

\def\sT{{\mathbb{T}}}

